\documentclass[letterpaper]{article} 
\usepackage{aaai23}  
\usepackage{times}  
\usepackage{helvet}  
\usepackage{courier}  
\usepackage[hyphens]{url}  
\usepackage{graphicx} 
\usepackage{natbib}  
\usepackage{caption} 
\usepackage{algorithm}
\usepackage{algorithmic}

\usepackage{newfloat}
\usepackage{listings}
\DeclareCaptionStyle{ruled}{labelfont=normalfont,labelsep=colon,strut=off} 
\floatstyle{ruled}
\newfloat{listing}{tb}{lst}{}
\floatname{listing}{Listing}
\usepackage{booktabs} 

\usepackage{amsmath}
\usepackage{amssymb}
\usepackage{mathtools}
\usepackage{amsthm}

\usepackage{subfigure}

\usepackage{graphicx}
\usepackage{float}
\usepackage{amsfonts,amsmath,amssymb,amsthm}
\usepackage{multicol, multirow}
\usepackage{makecell}
\usepackage[font=small]{caption}
\usepackage{adjustbox}
\usepackage{enumitem}

\usepackage{amsmath,amsfonts,bm}

\def\eqref#1{equation~\ref{#1}}

\def\1{\bm{1}}

\def\vh{{\bm{h}}}

\def\vm{{\bm{m}}}

\def\vx{{\bm{x}}}

\DeclareMathAlphabet{\mathsfit}{\encodingdefault}{\sfdefault}{m}{sl}
\SetMathAlphabet{\mathsfit}{bold}{\encodingdefault}{\sfdefault}{bx}{n}

\usepackage{caption}
\usepackage{cleveref}

\newcommand{\ie}{\em{i.e.}}
\newcommand{\eg}{\em{e.g.}}
\newcommand{\data}{RDB}
\newcommand{\method}{InfoNode}
\newcommand{\targetcol}{target column}  
\newcommand{\targettab}{target table}  
\newcommand{\targetnode}{target node}  
\newcommand{\rdbtograph}{\textsc{RDBToGraph}}  

\DeclareMathOperator{\Prob}{P}
\DeclareMathOperator{\Info}{H}
\DeclareMathOperator{\Mutual}{I}

\title{
Flaky Performances when Pretraining on Relational Databases
}
\author {
    Shengchao Liu,\footnotemark[1]\textsuperscript{\rm 1, \rm 2}
    David Vazquez,\textsuperscript{\rm 3}
    Jian Tang,\textsuperscript{\rm 1, \rm 4, \rm 5}
    Pierre-Andr\'e No\"el\textsuperscript{\rm 3}
}
\affiliations {
    \textsuperscript{\rm 1} Mila, 
    \textsuperscript{\rm 2} Université de Montréal,
    \textsuperscript{\rm 3} ServiceNow Research,
    \textsuperscript{\rm 4} HEC Montréal,
    \textsuperscript{\rm 5} CIFAR AI Chair\\
    liusheng@mila.quebec, pierre-andre.noel@servicenow.com
}

\usepackage{bibentry}

\begin{document}

\maketitle

\begin{abstract}
We explore the downstream task performances for graph neural network (GNN) self-supervised learning (SSL) methods trained on subgraphs extracted from relational databases (RDBs). Intuitively, this joint use of SSL and GNNs should allow to leverage more of the available data, which could translate to better results. However, we found that naively porting contrastive SSL techniques can cause ``negative transfer'': linear evaluation on fixed representations from a pretrained model performs worse than on representations from the randomly-initialized model. Based on the conjecture that contrastive SSL conflicts with the message passing layers of the GNN, we propose InfoNode: a contrastive loss aiming to maximize the mutual information between a node's initial- and final-layer representation. The primary empirical results support our conjecture and the effectiveness of InfoNode.
\end{abstract}

\section{Introduction}
The success story of large language models hinges on self-supervised learning (SSL). Deep neural networks (DNNs) in other domains have similarly benefited from different SSL techniques, including some based on image augmentations.

Yet relational database (RDB) are typically addressed by practitioners with fully-supervised, non-deep machine learning (ML) models: one first ``flattens'' the RDB to a single table, which enables the use of ML models accepting ``tabular data'', a domain that has recently been called ``the last \emph{unconquered castle} for deep learning'' \citep{kadra2021well}. An attack angle is that DNNs need not restrict themselves to tabular inputs: they may leverage more of the original RDB's structure, an hypothesis supported by graph neural networks (GNNs) work \cite{cvitkovic2020supervised}. However, publications on deep graph-based models for RDB data remain very rare: even with access to extra graph information, systematically beating tree models on {\data}s flattened with deep feature synthesis (DFS) remains a challenge \cite{cvitkovic2020supervised}.

SSL presents another opportunity for DNNs: we often have access to numerous unlabeled {\data} entries in addition to the few labeled ones. However, under certain circumstances, pretraining on unlabeled data (followed by linear probing) can perform worse than an untrained model, including some seemingly ``reasonable'' choices of SSL strategy (details in Appendix). Following these observations, we conjecture that the contrastive pretraining on RDB data is very sensitive to the view construction and to the backbone model. \Cref{sec:method} proposes \method, a contrastive method aiming to maximize the mutual information between a node's initial- and final-layer representation. \Cref{sec:experiment}'s primary empirical results show \method's effectiveness.

\textbf{Contributions.} To the best of our knowledge, we are the first to approach classification tasks on RDB using GNNs while leveraging unlabeled data using SSL pretraining. We empirically show that, while the use of SSL may confer some advantages for some datasets, SSL can actually lead to severe performance decrease, {\ie}, negative transfer. We introduce {\method}, which helps to certain extent.

\renewcommand{\thefootnote}{\fnsymbol{footnote}}
\footnotetext[1]{Work done during internship at ServiceNow Research.}

\begin{table*}[htb!]
\caption{\small
Main results with linear probing. In all cases, a linear classifier is trained on the representations of frozen models. For Untrained, the models are still in their randomly-initialized state. In the remaining rows, models are first pretrained with different SSL strategies before being frozen. Using \method{} alone may cause performances to drop; it's joint use with Generative (Hybrid) gives the overall better performances.
}
\label{tab:main_results}
\vspace{-2ex}
\label{tab:main_table}
\setlength{\tabcolsep}{5pt}
\fontsize{10}{6}\selectfont
\centering
\begin{adjustbox}{max width=\textwidth}
\begin{tabular}{c l c c c c c c c}
\toprule
\multirow{2}{*}{\makecell{SSL\\pretraining}} & \multirow{2}{*}{Model} & \multicolumn{2}{c}{Acquire (160k)} & \multicolumn{2}{c}{Home Credit (307k)} & \multicolumn{2}{c}{KDD Cup (619k)}\\
\cmidrule(lr){3-4} \cmidrule(lr){5-6} \cmidrule(lr){7-8}
& & S=10\% & S=100\% & S=10\% & S=100\% & S=10\% & S=100\% \\
\midrule
Untrained
& GCN & 54.36 $\pm$ 0.12 & 54.16 $\pm$ 0.22 & 52.00 $\pm$ 0.08 & 56.37 $\pm$ 1.72 & 51.78 $\pm$ 1.03 & 58.95 $\pm$ 0.45\\
(random init)
& PNA  & 58.71 $\pm$ 0.73 & 61.68 $\pm$ 0.33 & 55.75 $\pm$ 1.96 & 62.76 $\pm$ 0.96 & 56.08 $\pm$ 2.43 & 62.05 $\pm$ 1.29\\
\midrule
\multirow{2}{*}{Generative}
& GCN & 56.78 $\pm$ 0.08 & 58.19 $\pm$ 0.11 & 55.85 $\pm$ 0.00 & 62.52 $\pm$ 0.05 & 59.38 $\pm$ 0.05 & 64.38 $\pm$ 0.01\\
& PNA & 64.85 $\pm$ 0.12 & 66.34 $\pm$ 0.06 & \textbf{61.53 $\pm$ 0.04} & \textbf{67.43 $\pm$ 0.04} & \textbf{65.65 $\pm$ 0.01} & 69.11 $\pm$ 0.06\\
\cmidrule(lr){1-8}
\multirow{2}{*}{\makecell[t]{{\method}}}
& GCN & 53.76 $\pm$ 0.06 & 54.12 $\pm$ 0.09 & 54.78 $\pm$ 0.00 & 56.00 $\pm$ 0.04 & 52.69 $\pm$ 0.02 & 53.98 $\pm$ 0.08\\
& PNA & 51.51 $\pm$ 0.33 & 51.64 $\pm$ 0.22 & 55.09 $\pm$ 0.49 & 55.80 $\pm$ 0.40 & 51.81 $\pm$ 0.42 & 53.12 $\pm$ 0.12\\
\cmidrule(lr){1-8}
\multirow{2}{*}{Hybrid}
& GCN & 56.02 $\pm$ 0.15 & 58.19 $\pm$ 0.11 & 55.69 $\pm$ 0.02 & 59.79 $\pm$ 0.09 & 57.33 $\pm$ 0.01 & 60.21 $\pm$ 0.02\\
& PNA & \textbf{65.42 $\pm$ 0.00} & \textbf{66.60 $\pm$ 0.03} & 59.35 $\pm$ 0.07 & 66.46 $\pm$ 0.15 & 64.52 $\pm$ 0.07 & \textbf{70.24 $\pm$ 0.02}\\
\bottomrule
\end{tabular}
\end{adjustbox}
\end{table*}

\section{Self-supervised Learning on \data\, Graph \label{sec:method}} 
In this section, we discuss three SSL pretraining strategies---Generative, \method{} and Hybrid---on the RDB graph. In addition, we also observe that existing graph contrastive SSL methods can bring in severe negative transfer issue. Due to the page limit, we leave more discussions in the appendix.

\textbf{Generative SSL.}
Denoising tasks are one of the most widely-used generative SSL methods: here we mask-out a small fraction of the node attributes by replacing them by random values. Concretely, for each node $i$, a binary mask vector $\beta$ of the same length as $A_i$ is generated, a node $j$ of the same type as $i$ is randomly selected from the current batch, and the masked attributes are $A_i' = \beta \cdot A_i + (1-\beta) \cdot A_{j}$. The objective is then to recover the original attributes $A$ from the noisy representation $\vh'{}^T = \text{MPNN}(A', E)$. The loss $\mathcal{L}_G$ is a sum of mean squared errors for the continuous attributes and of cross entropies for the categorical ones.

\textbf{Contrastive SSL: InfoNode.}
``Over-smoothing'' is a well-known issue with GNNs: node-level representations may become indistinguishable and prediction performance may thus severely degrade as the number of layers increases. Conjecturing that it may be desirable for a node to ``remember about itself'', we introduce {\method}: a node's initial ($\vh_i^0$) and final ($\vh_i^T$) representations become two views for a contrastive loss ($\vh_g$ is the graph embedding):
{ \fontsize{7.8}{1}\selectfont
\begin{align} \label{eq:contrastive_02}
&\mathcal{L}_{\text{C-\method}} = \mathbb{E}_{(i, g) \in \text{Pos} } \big[ \sigma(f(\vh_i^0, \vh_i^T) + \sigma(f(\vh_i^T, \vh_g) \big] \\
& + \mathbb{E}_{(i', j') \in \text{Neg}} \big[ 1 - \sigma(f(\vh_{i'}^0, \vh_{j'}^T) \big] + \mathbb{E}_{(i', g') \in \text{Neg}} \big[ 1 - \sigma(f(\vh_{i'}^T, \vh_{g'}) \big]. \nonumber
\end{align}
}
\vspace{-2ex}

\textbf{Hybrid objective.}
We follow~\citet{liu2021pre}, where combining contrastive and generative SSL can augment the pretrained representation. 
Writing $\alpha_0$ and $\alpha_1$ the coefficients for the generative and contrastive objectives, the resulting objective function is:
\begin{equation}
\small{
\begin{aligned}
\mathcal{L} = \alpha_0 \cdot \mathcal{L}_{\text{G}} + \alpha_1 \cdot \mathcal{L}_{\text{C-\method}}.
\end{aligned}
}
\end{equation}

\section{Experiments and Discussion} \label{sec:experiment}
\textbf{Pipeline.} We adopt the pretraining and linear-probing pipeline, {\ie}, we will do SSL pretraining first, then we will fix the encoder and only fine-tune the prediction head. We adopt linear-probing because it can directly reflect the expressiveness of the pretrained model.

\textbf{Datasets and evaluation.} We consider the same 3 {\data} datasets as in~\citet{cvitkovic2020supervised}, all pre-processed with {\rdbtograph}. For these datasets, the predicted labels are binary and imbalanced, motivating the use of ROC-AUC. In addition, we use the whole training dataset for unsupervised pretraining, and then sample $S$\% for downstream.\looseness=-1

\textbf{Backbone models and baselines.}
For the backbone GNNs~\cite{kipf2016semi}, we consider 2 widely-used ones: GCN and PNA. The readout function is an attention module. For the pretraining methods, we first consider an untrained version ({\ie}, without any pretraining). Then we consider the generative SSL, contrastive SSL, and the hybrid of the two.

\textbf{Main results.}
\Cref{tab:main_results} reports linear probing (LP) results as an indicator of the quality of the representations learned by different SSL strategies. Generative SSL shows quite consistent improvements. Interestingly, contrastive methods taken on their own perform rather poorly from this linear probing perspective. The learned representations are at best comparable to random representations, and in many cases are much worse (``negative transfer''). While not being particularly impressive, the hybrid SSL results do not show this counter-intuitive behavior. This generative/contrastive dichotomy is less visible in fine tuning (Appendix), possibly because the models are given the opportunity to ``unlearn'' bad representations. This observation also holds for the other contrastive pretraining methods on graph, yet our proposed {\method} can alleviate the negative transfer issue better. Please see Appendix for more details.

\textbf{Analysis.}
According to our hypothesis, leveraging unlabeled data with SSL should typically improve downstream task performances. Of course, we were aware that there is no free lunch: due to its inductive biases, a model may be good for some tasks and bad for others. However, we believe that our results are not just random edge cases, but instead reveal a more systematic SSL failure mode. In particular, we posit that {\data} data distributions contain \emph{traps}---``interesting-looking noise''---that some SSL strategies may ``fall for'', and that ``better'' models may be more prone to these traps. As an illustration of how such traps may exist, consider a single-table {\data} with $3$ non-label columns---so a graph made of an isolated node with $3$ properties---and suppose that its probability distribution factorizes as  $\Prob(A) = \Prob(A_{00})\Prob(A_{01},A_{02})$. Given unlabeled data samples $A$, the ``best'' that any SSL strategy could do is to learn $\Prob(A_{00})$ and $\Prob(A_{01},A_{02})$.
The ability to uncover the presence of mutual information $\Mutual(A_{01};A_{02})$ between the corresponding datum is one of the characteristics typically associated with ``good'' SSL models, \emph{but such models may neglect $A_{00}$, and $A_{00}$ may be all that matters for some downstream tasks}. The Appendix further develops this conjecture.

\textbf{Conclusion and Future Direction.}
In this work, we propose a novel contrastive pretraining method, \method{}, to alleviate the inherent issue of GNN.
Primary experiments motivate certain insightful thinking. The mechanisms hypothesized could \textit{a priori} apply to both generative and contrastive SSL strategies. Investigating this dichotomy is part of our plan for a future characterization work.

\small{
\bibliography{reference}
}
\appendix
\section{Background}

\begin{figure*}[tb]
\vskip .1 in
\begin{center}
\centering%
\begin{subfigure}[{\data}.]%
{\includegraphics[width=0.47\linewidth]{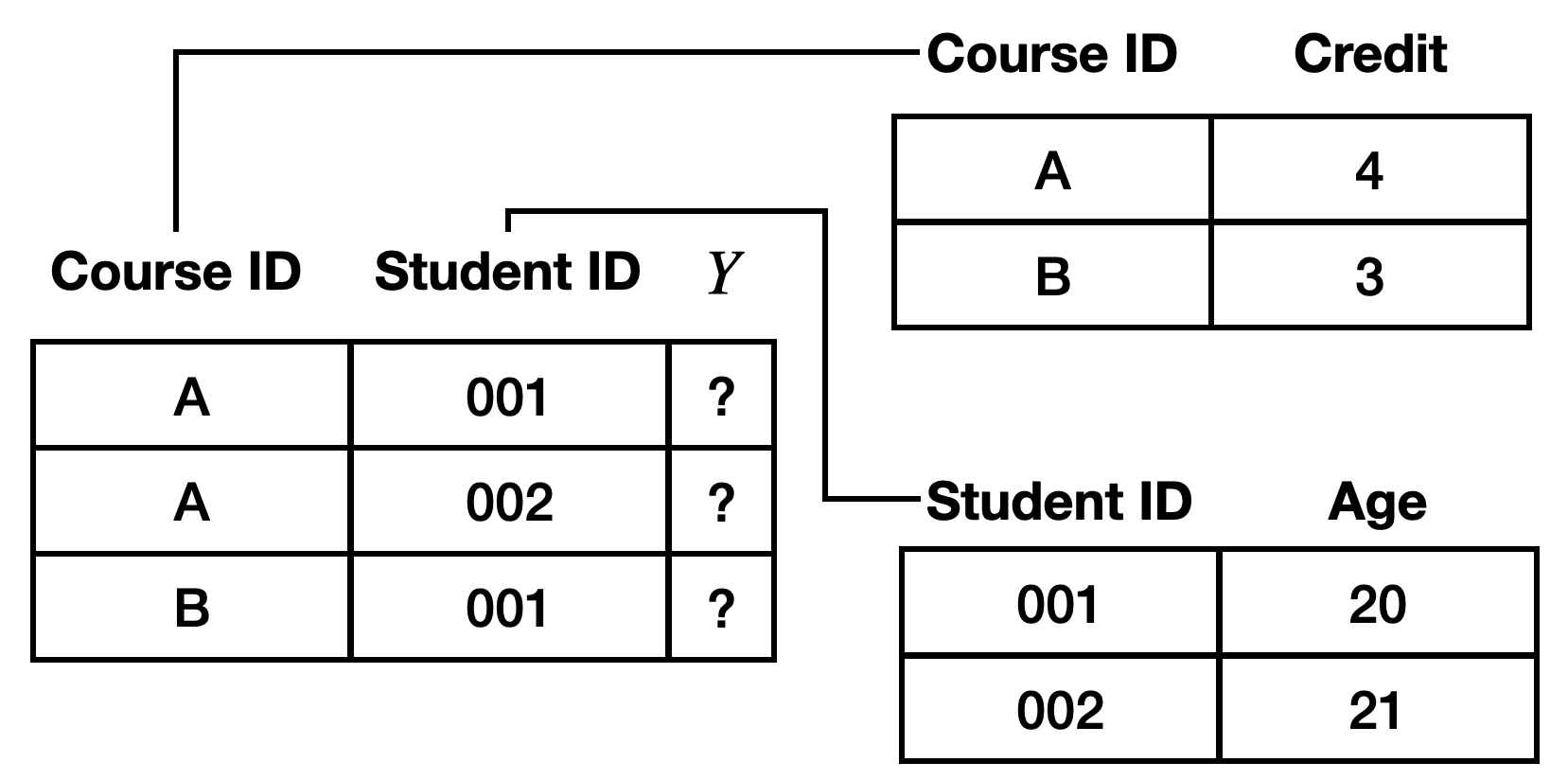}%
\label{fig:example:rdb}}%
\end{subfigure}%
\hfill%
\begin{subfigure}[Tabular data.]%
{\raisebox{\bigskipamount}{\includegraphics[width=0.47\linewidth]{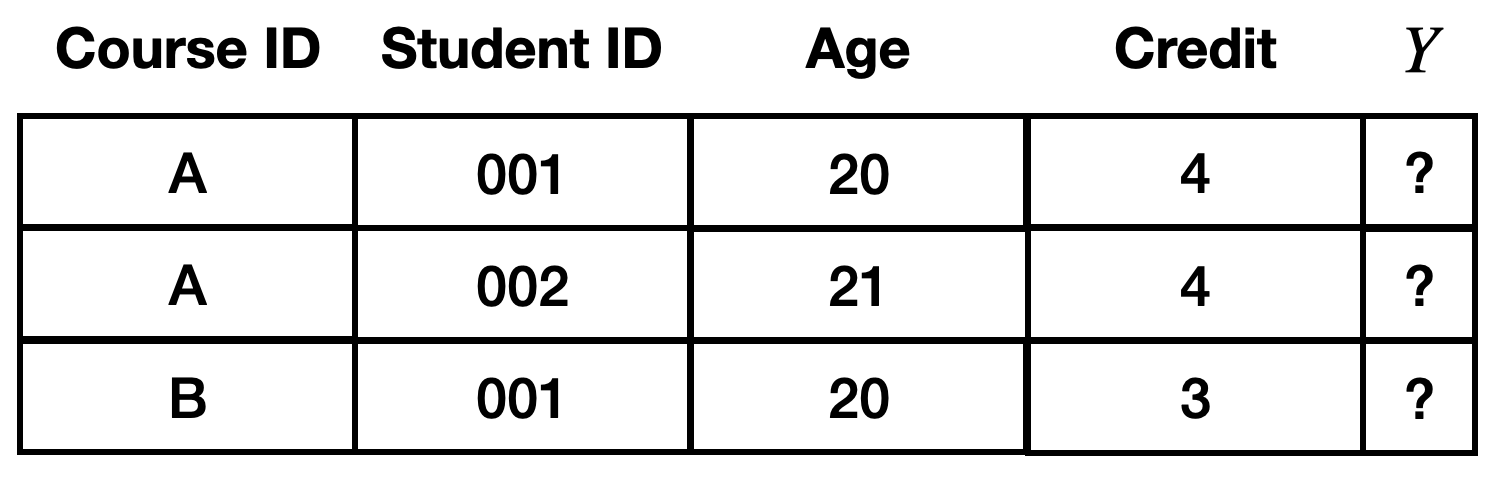}}%
\label{fig:example:tabular}}%
\end{subfigure}%
\caption{\small{\subref{fig:example:rdb} A relational database ({\data}) with three tables. \subref{fig:example:tabular}~The same information presented as tabular data (single table). Not all {\data}s may be so flattened without loss of information. In both cases, the task is to predict the value of the {\targetcol} $Y$.}}
\label{fig:example}
\end{center}
\vskip -.25 in
\end{figure*}

\textbf{Tabular data and relational databases.}
For our present purpose, we define a \emph{Relational Database} (RDB) as a collection of tables whose columns may reference one-another (or even themselves), while we reserve the term \emph{tabular data} to the case where there is a single table with no reference whatsoever. Figure~\ref{fig:example} gives examples of both these formats.

Following \citet{cvitkovic2020supervised}, we focus on the broad class of problems where the goal is to predict values from a specified \emph{\targetcol} of a specified \emph{\targettab} from a specified {\data}, given all available relevant information in the {\data}. 

However, it is common in applications that some (or even most) of the {\targetcol}'s entries are unavailable at training time ({\eg}, requiring human labelling and/or relating to future events), which precludes the use of the corresponding rows in a supervised learning paradigm. In this work, we extend the supervised learning problem statement \cite{cvitkovic2020supervised}, allowing algorithms to leverage such unlabeled data.

\textbf{Handling {\data}s in a Machine Learning Pipeline.} {\data}s (multiple tables) are commonly treated as tabular data (single table) in ML pipelines. Of course, we may seek to salvage some information from the other tables by adding columns to the {\targettab} through procedures such as join operations ({\eg}, taking Fig.~\ref{fig:example:rdb} to Fig.~\ref{fig:example:tabular}). Methods like DFS \cite{kanter2015deep} have been proposed to automate such feature engineering. Regardless of how the {\data} was converted to a single table, it may be fed to any machine learning algorithm for tabular data, including tree-based \cite{ho1995random,Chen:2016:XST:2939672.2939785} and deep learning \cite{arik2021tabnet,yoon2020vime,huang2020tabtransformer,somepalli2021saint,bahri2022scarf} ones.

Another approach represents a {\data} as an \emph{{\data} graph}---a graph with labeled directed edges, node types and node features that has the same informational content as the {\data}. Indeed, each row of a table can be represented as a node whose type identifies the corresponding table, the row's non-reference columns are stored as node features, and the row's reference columns specify the destination of a directed edge bearing a label identifying the corresponding column. The sampling strategy {\rdbtograph} \cite{cvitkovic2020supervised} has been proposed as a pre-processing step to generate one subgraph of the {\data} graph for each row of the {\targettab}.

Concretely, such an {\data} subgraph can be represented as the model input $\vx=(A,E)$. 
The node feature information $A$ is such that $A_i$ provides the values $A_{ik}$ for the $k$-th non-reference column of the row associated with node $i$ (excluding the {\targetcol} for the \emph{\targetnode}, {\ie}, the only node of this subgraph associated to the {\targettab}), to which we append the name of the associated table ({\ie}, the node's type). The edge information $E$ specifies the adjacency information: column $r$ of the row associated with node $i$ has a reference to the row associated with node $j$ if and only if $(i, r, j) \in E$ ({\ie}, there is a directed edge with label $r$ from node $i$ to node $j$).
When it is known, we note $y$ the entry in the {\targetcol} of the row corresponding to the {\targetnode} ({\ie}, the value to be predicted).

\textbf{Graph neural network on {\data} graphs.}
GNNs are DNNs designed so that their computational graph aligns with the data's graph structure, and many variations have been proposed \cite{kipf2016semi,gilmer2017neural,xu2018powerful,corso2020principal}. Given an input $\vx=(A,E)$, the first step is to embed the node feature information $A$ as an initial representation
\begin{equation}
\small{
\begin{aligned}
\vh_i^0 = \text{Embed}(A_i) \quad \forall i ,
\end{aligned}
}
\end{equation}
where the embedding function $\text{Embed}(\cdot)$ is typically adapted to each node's type.
The structural information $E$ is then leveraged to update this representation using an iterative scheme whose details depends on which kind of GNN is being used. In the case of message passing neural networks (MPNNs), message passing layers~\cite{gilmer2017neural} provide the node representation $\vh_i^t$ at layer $t$\looseness=-1
\begin{equation}
\small{
\begin{aligned}
  \vm_i^t & = \sum_{\mathclap{(i,r,j) \in E}} \, \text{Message}_{t-1}\left(\vh_i^{t-1}, r, \vh_j^{t-1}\right), ~~~\\
  \vh_i^t & = \text{Update}_{t-1}\left(\vh_i^{t-1}, \vm_i^t\right) ,
\end{aligned}
}
\end{equation}
where $\text{Message}_{t-1}$ and $\text{Update}_{t-1}$ are the message passing and node update functions on layer $t-1$, and $\vm_i^t$ is the message representation. Note that the message's dependency on $r$ is not actually used in this work\footnote{The datasets we consider are such that, given an edge $(i, r, j) \in E$, the type of node $i$ and the type of node $j$ uniquely specify the value of $r$, so having $\text{Message}_{t-1}$ depend on $r$ would be redundant information.}.

Repeating this for $T$ message passing layers provides the node-level representation $\vh_i^T$, which encodes information up to the $T$-hops neighborhood of node $i$. If a graph-level representation $\vh_g$ is required, it can be obtained by by applying a readout function $\text{Readout}$ to the aggregate $\vh^T$ of all the individual $\vh_i^T$
\begin{equation}
\small{
\begin{aligned}
\vh_g = \text{Readout}\left( \vh^T \right) .
\end{aligned}
}
\end{equation}
Finally, the prediction $\hat{y}$ for the {\targetcol} of the {\targettab} can be obtained by applying a multi-layer perceptron (MLP) to the graph-level representation $\vh_g$, which is the method favoured in \cite{cvitkovic2020supervised}. Alternatively, because of the special role played by the {\targetnode} in the {\data} graph, we may directly apply the MLP to the {\targetnode}'s representation $\vh_0^T$.

\section{Related Work}
Self-supervised learning (SSL) methods have attracted massive attention in graph structured data \citep{hu2019strategies,sun2019infograph,liu2019ngram,You2020GraphCL}. SSL strategies are often divided in two main categories \citep{liu2021graph,xie2021self,wu2021self,liu2021self}: generative and contrastive.

\textbf{Generative SSL.}
Generative SSL focuses on reconstructing the original sample at the \textbf{intra-data} level. For example, \citet{hu2019strategies} mask some nodes in the graph, and do the reconstruction on the masked items. More recently, \citet{liu2022molecular} add noise to the pairwise distances in the 3D molecular graph and the goal is to reconstruct the original distances.

\textbf{Contrastive SSL.}
Contrastive SSL gets its supervised signals from the \textbf{inter-data} level. Positive and negative view pairs are first defined, and the training task amounts to align the representations of positive pairs while contrasting the negative ones. How such view pairs are defined is highly flexible. For example, Infograph~\cite{sun2019infograph,velickovic2019deep} uses node-graph pairs from the same graph as positives and pairs from different graphs as negatives.

\section{Contrastive SSL}
In this section, we will introduce InfoGraph~\cite{sun2019infograph}, a widely-used graph contrastive learning method.

In general, contrastive SSL maximizes the mutual information between two views of the data. Taking the two views to be the node- and graph-level representations has been widely explored, including InfoGraph. Concretely, a pair of node- and graph-level representations is positive if the node comes from that graph and negative if it comes from another graph of the batch. Given a function $f(\cdot,\cdot)$ (here the dot product) and noting $\text{Pos}$ (resp.\ $\text{Neg}$) the set of positive (resp.\ negative) pairs, the EBM-NCE objective \cite{liu2021pre} is:
\begin{equation} \label{eq:contrastive_01}
\small{
\begin{aligned}
\mathcal{L}_{\text{C-InfoGraph}}
    & = \mathbb{E}_{(i, g) \in \text{Pos} } \big[ \sigma(f(\vh_i^T, \vh_g) \big] \\
    & ~~~ + \mathbb{E}_{(i', g') \in \text{Neg}} \big[ 1 - \sigma(f(\vh_{i'}^T, \vh_{g'}) \big].
\end{aligned}
}
\end{equation}

\section{Linear Probing Results \label{appendix:linear_probing}}

Table~\ref{tab:app:main_results} reports linear probing (LP) results as an indicator of the quality of the representations learned by different SSL strategies. Interestingly, contrastive methods taken on their own perform rather poorly from this linear probing perspective: the learned representations are at best comparable to random representations, and in many cases are much worse (``negative transfer''). While not being particularly impressive, generative and hybrid SSL results do not show this counter-intuitive behavior. Our understanding is that this generative/contrastive dichotomy is less visible in Table~\ref{tab:app:main_results} because the models are given the opportunity to ``unlearn'' bad representations during fine-tuning. The mechanisms hypothesized in \Cref{figure:plan} could \textit{a priori} apply to both generative and contrastive SSL strategies, and investigating the mechanistic source of this dichotomy is part of our plan for a future characterization work.

\begin{table*}[htb!]
\caption{\small{
Main results with linear probing. In all cases, a linear classifier is trained on the representations of frozen models. For Untrained, these models are still in their randomly-initialized state. In the remaining rows, the models were first pretrained with the different SSL strategies before being frozen. Using \method{} alone may cause performances to drop, it's joint use with Generative (Hybrid) gives the overall better performances.
}}
\label{tab:app:main_results}
\vspace{-2ex}
\setlength{\tabcolsep}{5pt}
\fontsize{10}{6}\selectfont
\centering
\begin{adjustbox}{max width=\textwidth}
\begin{tabular}{c l c c c c c c c}
\toprule
\multirow{2}{*}{\makecell{SSL\\pre-training}} & \multirow{2}{*}{Model} & \multicolumn{2}{c}{Acquire (160k)} & \multicolumn{2}{c}{Home Credit (307k)} & \multicolumn{2}{c}{KDD Cup (619k)}\\
\cmidrule(lr){3-4} \cmidrule(lr){5-6} \cmidrule(lr){7-8}
& & S=10\% & S=100\% & S=10\% & S=100\% & S=10\% & S=100\% \\
\midrule
Untrained
& GCN & 54.36 $\pm$ 0.12 & 54.16 $\pm$ 0.22 & 52.00 $\pm$ 0.08 & 56.37 $\pm$ 1.72 & 51.78 $\pm$ 1.03 & 58.95 $\pm$ 0.45\\
(random init)
& PNA  & 58.71 $\pm$ 0.73 & 61.68 $\pm$ 0.33 & 55.75 $\pm$ 1.96 & 62.76 $\pm$ 0.96 & 56.08 $\pm$ 2.43 & 62.05 $\pm$ 1.29\\
\midrule
\multirow{2}{*}{Generative}
& GCN & 56.78 $\pm$ 0.08 & 58.19 $\pm$ 0.11 & 55.85 $\pm$ 0.00 & 62.52 $\pm$ 0.05 & 59.38 $\pm$ 0.05 & 64.38 $\pm$ 0.01\\
& PNA & 64.85 $\pm$ 0.12 & 66.34 $\pm$ 0.06 & 61.53 $\pm$ 0.04 & 67.43 $\pm$ 0.04 & 65.65 $\pm$ 0.01 & 69.11 $\pm$ 0.06\\
\cmidrule(lr){1-8}
Contrastive:
& GCN & 53.97 $\pm$ 0.26 & 54.52 $\pm$ 0.53 & 52.75 $\pm$ 0.07 & 55.54 $\pm$ 0.06 & 51.59 $\pm$ 0.00 & 52.05 $\pm$ 0.04\\
InfoGraph:
& PNA & 50.83 $\pm$ 0.24 & 54.98 $\pm$ 0.22 & 52.68 $\pm$ 0.30 & 54.50 $\pm$ 0.69 & 51.20 $\pm$ 0.34 & 51.44 $\pm$ 0.10\\
\cmidrule(lr){1-8}
Hybrid:
& GCN & 57.47 $\pm$ 0.13 & 58.43 $\pm$ 0.09 & 54.20 $\pm$ 0.01 & 60.54 $\pm$ 0.04 & 59.00 $\pm$ 0.01 & 62.86 $\pm$ 0.03\\
InfoGraph:
& PNA & 63.78 $\pm$ 0.08 & 66.48 $\pm$ 0.02 & 59.57 $\pm$ 0.15 & 67.13 $\pm$ 0.14 & 55.39 $\pm$ 0.05 & 66.18 $\pm$ 0.02\\
\cmidrule(lr){1-8}
Contrastive:
& GCN & 53.76 $\pm$ 0.06 & 54.12 $\pm$ 0.09 & 54.78 $\pm$ 0.00 & 56.00 $\pm$ 0.04 & 52.69 $\pm$ 0.02 & 53.98 $\pm$ 0.08\\
\method
& PNA & 51.51 $\pm$ 0.33 & 51.64 $\pm$ 0.22 & 55.09 $\pm$ 0.49 & 55.80 $\pm$ 0.40 & 51.81 $\pm$ 0.42 & 53.12 $\pm$ 0.12\\
\cmidrule(lr){1-8}
hybrid:
& GCN & 56.02 $\pm$ 0.15 & 58.19 $\pm$ 0.11 & 55.69 $\pm$ 0.02 & 59.79 $\pm$ 0.09 & 57.33 $\pm$ 0.01 & 60.21 $\pm$ 0.02\\
\method
& PNA & 65.42 $\pm$ 0.00 & 66.60 $\pm$ 0.03 & 59.35 $\pm$ 0.07 & 66.46 $\pm$ 0.15 & 64.52 $\pm$ 0.07 & 70.24 $\pm$ 0.02\\
\bottomrule
\end{tabular}
\end{adjustbox}
\end{table*}

\section{Updated hypothesis and future work.\label{sec:hypothesis}}
\begin{figure}[tb!]
\vskip 0.1in
\begin{center}
\centerline{\includegraphics[width=0.9\columnwidth]{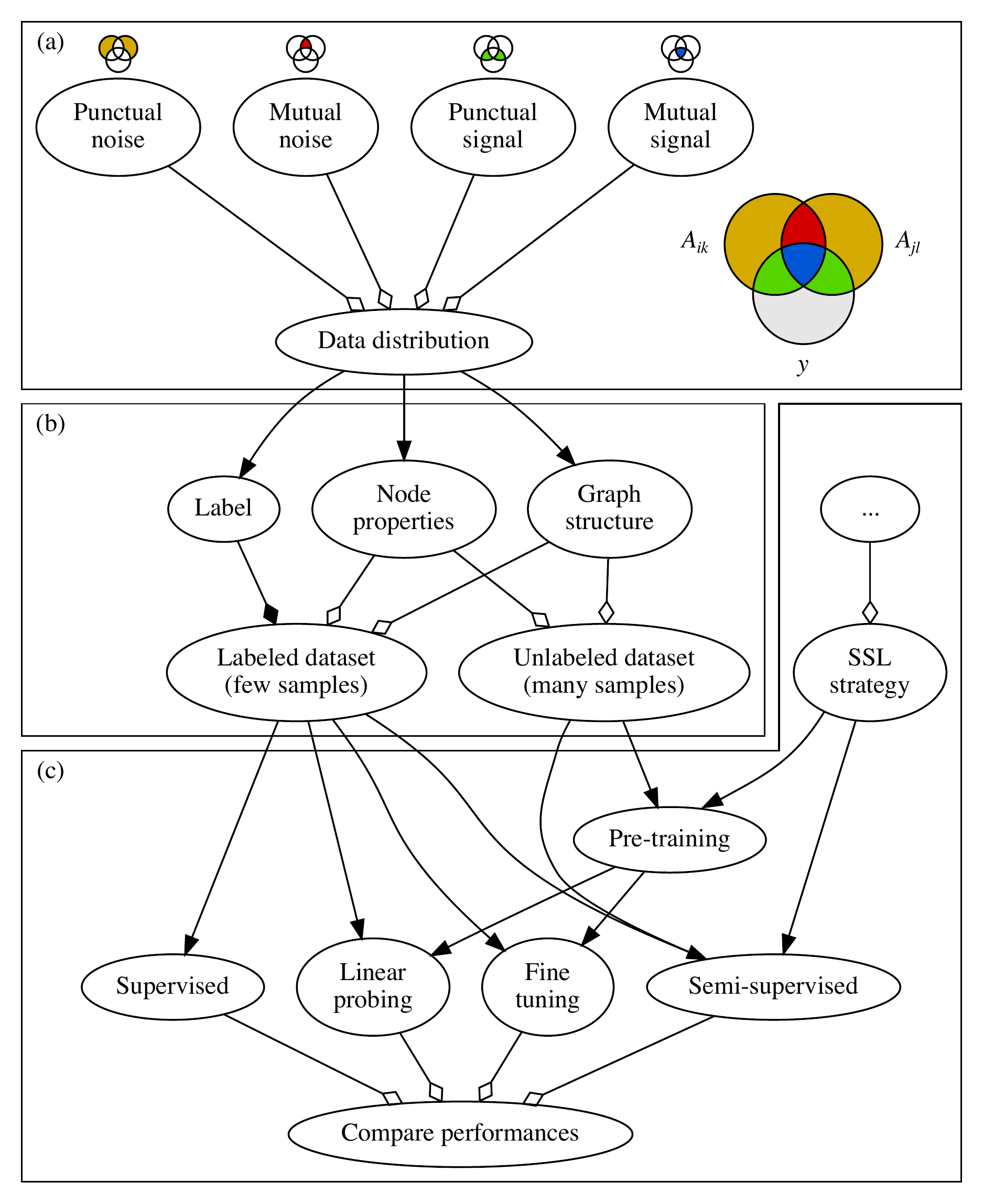}}
\vspace{-2ex}
\caption{\small{Plan for a future characterization work.
Arrows indicate information flow while diamonds represent ``and/or'' aggregation (closed diamond means ``always present'').
\textbf{(a)}
We distinguish four kinds of contributions to the data distribution using an information-theoretic perspective.
We call \emph{punctual noise} the information from an entry $A_{ik}$ that is independent from both the label $y$ and the other entries $A \setminus \{A_{ik}\}$, and \emph{punctual signal} the information independent from other entries but shared with $y$.
Similarly, we call \emph{mutual noise} the information shared by two (or more) entries $A_{ik}$ and $A_{jl}$ but not with $y$, and \emph{mutual signal} the one shared with $y$.
\textbf{(b)}
Conceptually, we can understand (labeled and unlabeled) datasets as being ``sampled'' from the data distribution.
How much of the node properties and graph structure actually ``makes it'' to the model depends on the pre-processing and representation strategies: using the {\targettab} on its own gets less information than DFS \cite{kanter2015deep}, which itself gets less than {\rdbtograph} \cite{cvitkovic2020supervised}.
But ``more information'' does not necessarily imply better downstream task performances: part of that information is ``noise''.
\textbf{(c)}
While supervised learning may directly learn to ignore ``noise'' contributions and focus on ``signal'' ones, only ``punctual'' and ``mutual'' information may be distinguished using unlabeled data alone. Our plan is to obtain analytical bounds and measure empirical performances for different training paradigms and/or SSL strategies, for real and synthetic datasets.
}}
\label{figure:plan}
\end{center}
\vskip -0.25in
\end{figure}

According to our original hypothesis, leveraging unlabeled data with SSL should typically improve downstream task performances. Of course, we were aware that there is no free lunch \citep{wolpert1997no}: due to its inductive biases, a model may be good for some tasks and bad for others. However, we believe that our results are not just random edge cases, but instead reveal a more systematic SSL failure mode. In particular, we posit that {\data} data distributions contain \emph{traps}---``interesting-looking noise''---that some SSL strategies may ``fall for'', and that ``better'' models may be more prone to these traps.

\textbf{Example of a simple ``trap''.}
As an illustration of how such traps may exist, consider a single-table {\data} with $3$ non-label columns---so a graph made of an isolated node with $3$ properties---and suppose that its probability distribution factorizes as  $\Prob(A) = \Prob(A_{00})\Prob(A_{01},A_{02})$. Given unlabeled data samples $A$, the ``best'' that any SSL strategy could do is to learn $\Prob(A_{00})$ and $\Prob(A_{01},A_{02})$. The ability to uncover the presence of mutual information $\Mutual(A_{01};A_{02})$ between the corresponding datum is one of the characteristics typically associated with ``good'' SSL models.

Now further suppose that $\Prob(y|A) = \Prob(y|A_{00})$: any model capacity dedicated on learning $\Prob(A_{01}, A_{02})$ during pre-training has, in retrospect, been wasted. From the perspective of predicting $y$, $A_{01}$ and $A_{02}$ are ``noise'', all the ``signal'' resides in $A_{00}$. ``Better'' models, capable of uncovering the intricate dependences in $\Prob(A_{01},A_{02})$, are more prone to fall for the trap of noise made ``interesting-looking'' by its mutual information.

\textbf{Data distribution.}
More generally, we concretize the noise/signal and mutual/punctual dichotomies alluded to above and distinguish four kinds of contributions to the data distribution, illustrated in \cref{figure:plan}(a).
In each case, we quantify the information involved for a pair of datum $(A_{ik}, A_{jl})$ in relation to the label $y$, but this is only an example: our proposed naming convention is meant to generalize beyond such pairs \citep{bell2003co}, as reflected in the descriptive text.
\begin{itemize}
\item \emph{Punctual noise}, \textit{e.g.}, $\Info(A_{ik}|A_{jl},y)+\Info(A_{jl}|A_{ik},y)$, is the information in individual datum that is independent from both the label and the rest of the input.
\item \emph{Punctual signal}, \textit{e.g.}, $\Mutual(A_{ik};y|A_{jl}) + \Mutual(A_{jl};y|A_{ik})$, is the mutual information between individual datum and the label, but independent from the rest of the input.
\item \emph{Mutual noise}, \textit{e.g.}, $\Mutual(A_{ik};A_{jl}|y)$, is the information that is shared by more than one datum, but that is independent of the label.
\item \emph{Mutual signal}, \textit{e.g.}, $\Mutual(A_{ik};A_{jl}) - \Mutual(A_{ik};A_{jl}|y)$, is the information that is shared by more than one datum as well as by the label. It is a co-information \citep{bell2003co}, \textit{e.g.}, $\Mutual(A_{ik};A_{jl};y)$, and may thus be negative.%
\footnote{%
For example, suppose that $A_{ik}$ and $A_{jl}$ are binary coin flips and that $y = A_{ik} \,\textup{XOR}\, A_{jl}$. There are only $2$ bits to be known about $(A_{ik},A_{jl},y)$, but each of those three quantities independently has $1$ bit of entropy. No pair among those three quantities has nonzero mutual information, but the co-information of the three of them together is $-1$ bit. Interestingly, ``good'' SSL pre-training strategies would conclude that there is likely nothing to be gained by considering $A_{ik}$ and $A_{jl}$ together, whereas their joint knowledge actually gives a perfect predictor of $y$.
}
\end{itemize}

Notice that what constitutes ``noise'' or ``signal'' is here explicitly dependent on the downstream task: it is \textit{a priori} impossible to distinguish noise from signal at pre-training time. However, it is possible to distinguish ``mutual'' contributions from ``punctual'' ones: many SSL pre-training strategies have the built-in inductive bias that ``mutual is a predictor for signal''. In those cases, systematic failure modes may be associated with a strong mutual noise and/or punctual signal, both present in the above ``trap'' example.

\textbf{Dataset pre-processing and model input.}
Models with such a ``mutual is a predictor for signal'' bias may not fall for traps they can't see, making pre-processing and input representation an important aspect of understanding how this inductive bias may affects pre-training performances (\cref{figure:plan}(b)). In its simplest form, this is rather trivial: the model cannot learn from columns and/or tables that were dropped during the pre-processing of a {\data}. But things may be more subtle: a model's ability to leverage graph structure may play against it.

Suppose that there are $N$ nodes, each with a single property $A_{i0}$. Further suppose that neighbouring nodes $(i,j)$ exhibit mutual noise $\Mutual(A_{i0},A_{j0}|y)$, and that there is otherwise no mutual information in $A$. If the number of edges is much smaller than $\frac{1}{2}N(N-1)$, a model that has no access to the graph structure may be unable to identify such mutual information. Conversely, a graph-aware model may have an inductive bias that neighbouring nodes are more likely to exhibit mutual information, and thus fall for the trap.

\textbf{Future characterization work.}
We posited that many SSL strategies may share the ``mutual is a predictor for signal'' inductive bias, and that this could contribute to why pre-training may cause negative transfer. But this mutual/punctual dichotomy is definitely not the sole angle we could consider and---far from having established cause-effect relationships---we merely glossed over the plausibility of such mechanisms.

\Cref{figure:plan} shows an high-level plan to address such issues in a future work.
Our goal is to obtain analytical bounds and measure empirical performances for different training paradigms and/or SSL strategies, probing them using datasets as the independent variable. To this end, we will characterize real-world datasets and specify synthetic ones in terms of different high-level properties, such as the four kinds of contributions listed above as well as their relation to the graph structure.

In light of recent results indicating that pre-training molecular graphs may also cause negative transfer \cite{wang2022evaluating,liu2022molecular}, it is natural to wonder what are the commonalities shared with {\data}. The graph nature of the data is an obvious suspect, but it is not clear if it is part of the root cause, if it merely exacerbates an already-present problem, and/or if the problem is actually inherent to the tool---GNNs---and not to the data.

\end{document}